\documentclass[10pt,twocolumn,letterpaper]{article}

\usepackage[pagenumbers]{iccv} 

\newcommand{\red}[1]{{\color{red}#1}}
\newcommand{\blue}[1]{{\color{blue}#1}}

\usepackage[hypcap=false]{caption}

\definecolor{iccvblue}{rgb}{0.21,0.49,0.74}
\usepackage[pagebackref,breaklinks,colorlinks,allcolors=iccvblue]{hyperref}

\usepackage{soul}

\usepackage{bm}
\usepackage{multirow}
\usepackage{stfloats}
\usepackage{marvosym}
\usepackage{pifont}

\def\paperID{12510} 
\def\confName{ICCV}
\def\confYear{2025}

\title{PixelHacker: Image Inpainting with Structural and Semantic Consistency}

\author{
Ziyang Xu\textsuperscript{\rm 1, $\diamond$} \quad
Kangsheng Duan\textsuperscript{\rm 1, $\diamond$} \quad
Xiaolei Shen\textsuperscript{\rm 2} \quad
Zhifeng Ding\textsuperscript{\rm 2} \quad
Wenyu Liu\textsuperscript{\rm 1}
\\
Xiaohu Ruan\textsuperscript{\rm 2} \quad
Xiaoxin Chen\textsuperscript{\rm 2} \quad
Xinggang Wang\textsuperscript{\rm 1, \Letter}
\vspace{0.3em}
\\
\textsuperscript{\rm 1}Huazhong University of Science and Technology \quad \textsuperscript{\rm 2}VIVO AI Lab
\\
}

\begin{document}
\maketitle

\makeatletter\def\Hy@Warning#1{}\makeatother
\let\thefootnote\relax\footnotetext{\textsuperscript{$\diamond$} Intern of VIVO AI Lab.}
\let\thefootnote\relax\footnotetext{\textsuperscript{\Letter} Corresponding Author:xgwang@hust.edu.cn}

\begin{abstract}
Image inpainting is a fundamental research area between image editing and image generation. Recent state-of-the-art (SOTA) methods have explored novel attention mechanisms, lightweight architectures, and context-aware modeling, demonstrating impressive performance. However, they often struggle with complex structure (e.g., texture, shape, spatial relations) and semantics (e.g., color consistency, object restoration, and logical correctness), leading to artifacts and inappropriate generation. To address this challenge, we design a simple yet effective inpainting paradigm called latent categories guidance, and further propose a diffusion-based model named PixelHacker. Specifically, we first construct a large dataset containing 14 million image-mask pairs by annotating foreground and background (potential 116 and 21 categories, respectively). Then, we encode potential foreground and background representations separately through two fixed-size embeddings, and intermittently inject these features into the denoising process via linear attention. Finally, by pre-training on our dataset and fine-tuning on open-source benchmarks, we obtain PixelHacker. Extensive experiments show that PixelHacker comprehensively outperforms the SOTA on a wide range of datasets (Places2, CelebA-HQ, and FFHQ) and exhibits remarkable consistency in both structure and semantics. Project page at \href{https://hustvl.github.io/PixelHacker}{https://hustvl.github.io/PixelHacker}.
\end{abstract}

\section{Introduction}
\label{sec:intro}

Image inpainting, as a fundamental research in computer vision, has been widely applied in image editing and object removal \cite{Sargsyan2023migan,wang2023imagen}. The goal is to generate visually plausible content within the masked region by leveraging contextual pixel information from a given image-mask pair \cite{liu2018partialinpainting}.

With the widespread application of inpainting techniques \cite{Bertalmio2000imageinpaint}, there is an increasing demand for higher visual quality in generated content, particularly in terms of structural and semantic consistency. Structural consistency ensures that the generated content exhibits natural texture transitions, shape coherence, and physically plausible spatial relationships with the surrounding pixels. Semantic consistency requires smooth color blending, faithful reconstruction of object features when the mask partially covers a target, and logical coherence within the contextual pixel environment. Recent state-of-the-
art (SOTA) approaches have focused on novel attention mechanisms, lightweight architectures, and improved contextual awareness, demonstrating impressive inpainting capabilities \cite{li2022mat,Sargsyan2023migan,suvorov2021lama}. However, they often struggle with complex structures and semantics, as shown in Fig.~\ref{fig:cover_showcase}.
For instance, in Fig.~\ref{fig:cover_showcase} (a), when dealing with a texture-rich tree trunk and a sunlit water surface, these methods produce semantic inconsistencies (e.g., generating irrelevant objects on the tree trunk), structural distortions, color discrepancies, and blurriness. Similarly, in Fig.~\ref{fig:cover_showcase}(b), when handling an image with multiple foreground elements (ground and people), middle-ground elements (railings and tree trunk), and background elements (forest and pathway), these methods fail to maintain structural consistency (e.g., discontinuous railings behind people), logical coherence, semantic relevance (e.g., generating objects unrelated to the scene), and also blurriness. 

To address these challenges, we design a simple yet effective inpainting paradigm, termed Latent Categories Guidance (LCG), and further propose a diffusion-based model named PixelHacker. Specifically, we first construct a large-scale dataset containing 14 million image-mask pairs by annotating “foreground” and “background” (potential 116 and 21 categories, respectively). Then, we employ two fixed-size embeddings to encode the latent foreground and background representations separately. By intermittently injecting these latent features into the denoising process via linear attention, we effectively guide the generation process toward structural and semantic interaction. Finally, by pre-training on our dataset and fine-tuning on open-source benchmarks, we obtain PixelHacker, a model that exhibits surprisingly well structural and semantics consistency.
As illustrated in Fig.~\ref{fig:cover_showcase}, PixelHacker generates visually coherent content with natural texture transitions, smooth color blending, and logical consistency, significantly outperforming existing SOTA methods. Notably, during the entire training pipeline, we do not require explicit supervision on the exact object category within the masked region (e.g., distinguishing between a person, car, or chair). Instead, we represent each masked object as either foreground or background, which encourages focusing on foreground-background semantics, implicitly compressing diverse object representations while reducing implementation costs.
Extensive experiments demonstrate that PixelHacker consistently outperforms SOTA methods across various benchmarks (Places2 \cite{zhou2017places}, CelebA-HQ \cite{karras2018celebahq}, and FFHQ \cite{karras2018stylegan_ffhq}), exhibiting superior structural and semantic consistency.

In summary, we make the following contributions:

(1) We introduce a simple yet effective inpainting paradigm, Latent Categories Guidance (LCG). LCG first constructs image-mask pairs based on “foreground” and “background” label. During training, LCG uses two fixed-size embeddings to encode the latent foreground and background features, which are intermittently injected into the denoising process via linear attention, leading to a model that inpaints with structural and semantic consistency.

(2) We propose PixelHacker, a diffusion-based inpainting model trained with LCG on 14M image-mask pairs and fine-tuning on open-source benchmarks. We demonstrate that PixelHacker achieves excellent structural and semantic consistency, outperforming various SOTA methods.

(3) Extensive experiments show that PixelHacker consistently outperforms SOTA approaches across multiple benchmarks (Places2, CelebA-HQ, and FFHQ).

\begin{figure*}[t]
\centering
\includegraphics[width=1.0\linewidth]{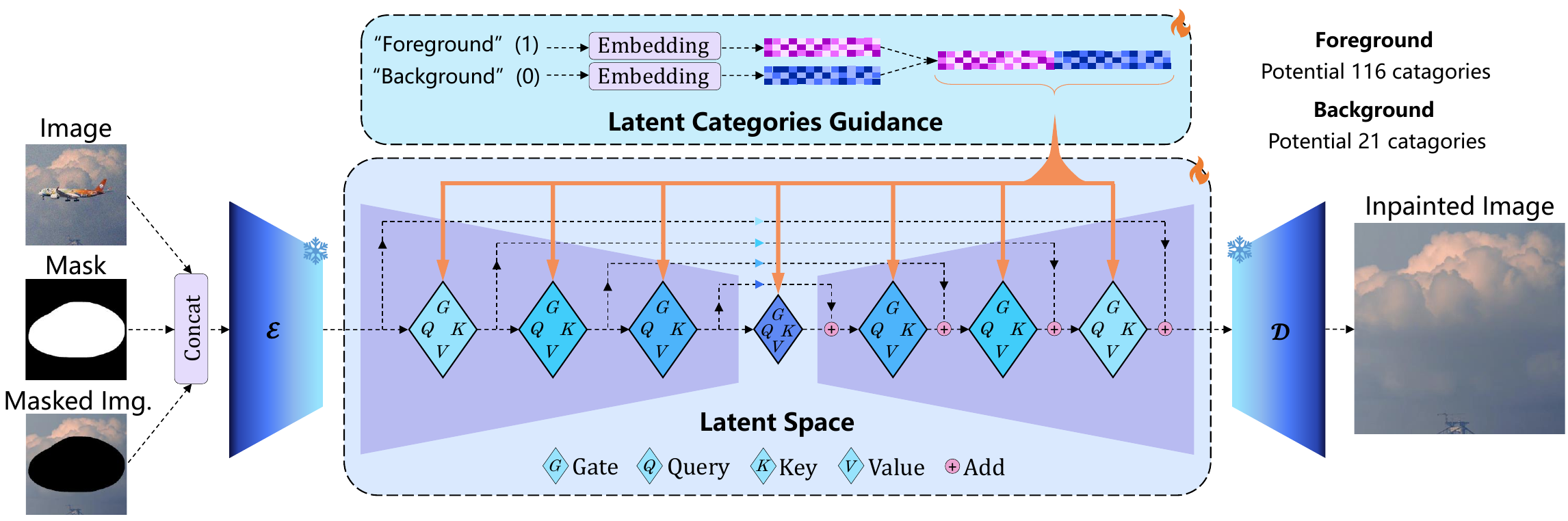}
\caption{\textbf{Overall pipeline of our PixelHacker.} PixelHacker builds upon the latent diffusion architecture by introducing two fixed-size LCG embeddings to separately encode latent foreground and background features. We employ linear attention to inject these latent features into the denoising process, enabling intermittent structural and semantic multiple interactions. This design encourages the model to learn a data distribution that is both structurally and semantically consistent. We elaborate on the interaction details in Fig.~\ref{fig:interaction} and Sec.~\ref{sec:interaction}.}
\label{fig:pipeline}
\end{figure*}

\section{Related Work}
\label{sec:relate}

\paragraph{Image Inpainting.}
GAN-based methods \cite{xu2023shgan, zhao2021comodgan, Sargsyan2023migan, karras2018stylegan_ffhq, Karras2019stylegan2, zeng2023AOTGAN, Choi2018stargan, Choi2020starganv2} typically introduce a large number of random masks during training to adversarially optimize the generator, enabling effective utilization of contextual texture features and achieving reasonable structural consistency. However, these methods often fail to infer semantic information from the masked region, leading to semantically implausible objects, as exemplified by MI-GAN in Fig.~\ref{fig:cover_showcase}(b).
Convolution-based approaches \cite{liu2018partialinpainting, yu2018deepfillv1, yu2019deepfillv2, li2022mat, suvorov2021lama} focus primarily on contextual coherence. While they expand the receptive field and mitigate the interference of irrelevant features, they lack explicit attention to semantic consistency, often resulting in abrupt color transitions and inappropriate feature reconstructions, as observed in the results of LaMa and MAT in Fig.~\ref{fig:cover_showcase}(a).
Diffusion-based methods benefit from high-quality pretrained text-to-image (T2I) models, allowing them to inject new semantic information via text prompts \cite{Rombach2022LDM,podell2023sdxl, zhuang2023powerpaint, xie2023smartbrush, wang2023imagen}. However, text prompts act as a double-edged sword in inpainting. On one hand, low-quality prompts degrade inpainting performance; on the other hand, introducing multiple text encoders to enhance prompt quality \cite{podell2023sdxl} results in computational redundancy. While text encoders improve semantic-aware generation, they often compromise structural consistency. For instance, as shown in Fig.~\ref{fig:cover_showcase}(b), the results of SD-Inpainting \cite{Rombach2022LDM}, SDXL-Inpainting \cite{podell2023sdxl}, and PowerPaint \cite{zhuang2023powerpaint} exhibit discontinuous railings behind the people, which disrupts the original scene structure.
Unlike these methods, PixelHacker builds upon the latent diffusion architecture by introducing two fixed-size embeddings to separately encode latent foreground and background features. We employ linear attention to inject these latent features into the denoising process, enabling intermittent structural and semantic multiple interactions. This design encourages the model to learn a data distribution that is both structurally and semantically consistent.
Similar to other Latent Diffusion Models (LDMs), we utilize Classifier-Free Guidance (CFG) inference \cite{Ho2021CFG}. Ultimately, PixelHacker achieves SOTA performance across multiple benchmarks \cite{karras2018stylegan_ffhq, zhou2017places, karras2018celebahq}.

\section{Method}
\label{sec:method}

\begin{figure}
\centering
\includegraphics[width=1.0\linewidth]{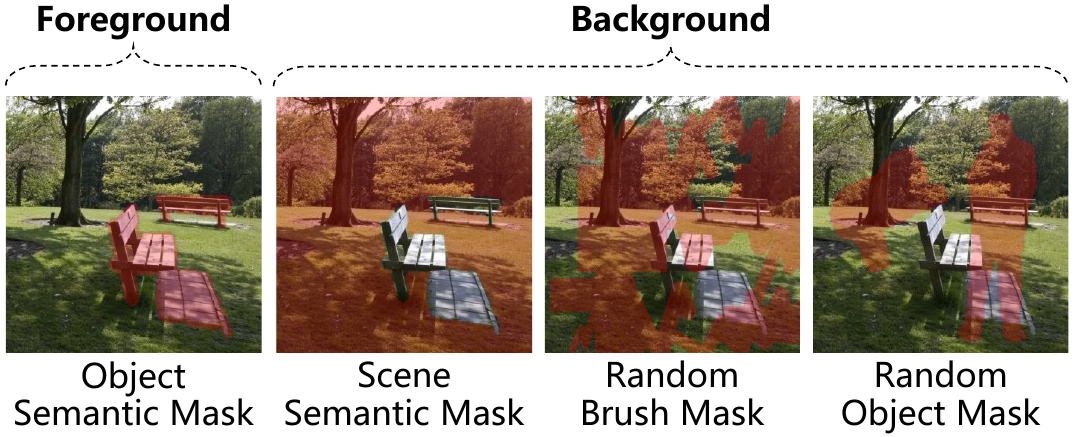}
\caption{
\textbf{Illustration of various masks we use to construct Latent Categories Guidance (LCG).} We assign object semantic masks to the foreground embedding and the other three masks to the background embedding. Details refer to Sec.~\ref{sec:CLCG}.
} 
\label{fig:CLCG}
\end{figure}

\subsection{Overall Pipeline}

The overall pipeline of our PixelHacker is illustrated in Fig.~\ref{fig:pipeline}. First, following \cite{Rombach2022LDM}, PixelHacker takes a noised image, clean mask, and clean masked image as inputs, concatenates them, and feeds them into the encoder of a VAE \cite{kingma2022VAE}, which transforms features from the pixel space to the latent space.
Next, LCG constructs image-mask pairs based on “foreground” and “background” label (potential 116 and 21 categories, respectively). Two fixed-size embeddings are then used to encode the latent foreground and background features separately.
Then, in the latent space, we apply linear attention throughout both the downsampling and upsampling processes. By injecting the embeddings into the linear attention, we achieve intermittent structural and semantic consistency interactions.
Finally, the encoded features in the latent space are passed through the decoder of the VAE to reconstruct the inpainted image.

\subsection{Construction of Latent Categories Guidance}
\label{sec:CLCG}

Previous studies have demonstrated that inpainting paradigms trained with random brush masks can achieve remarkable performance \cite{Pathak2016CtxtEnc, yu2018deepfillv1, Rombach2022LDM}. However, this strategy does not leverage segmentation masks, which contain rich semantic information.
On the other hand, while we enumerate 116 foreground categories and 21 background categories when constructing the dataset, the fixed number of categories inevitably limits the model's generalization and expansion capability for novel categories \cite{Dhariwal2021CG}.

\begin{figure}[t]
\centering
\includegraphics[width=1.0\linewidth]{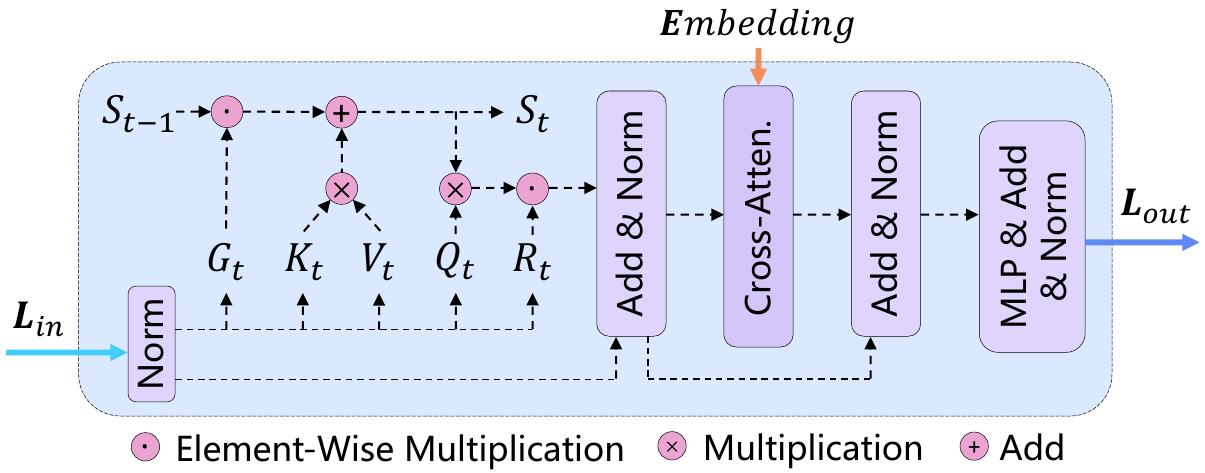}
\caption{\textbf{The single interaction process between LCG embeddings and latent features.} We elaborate on the details in Sec.~\ref{sec:interaction}. Throughout the pipeline, multiple interactions are performed in a sequential manner, guiding the model to learn foreground semantics, background semantics, and contextual structures.}
\label{fig:interaction}
\end{figure}

To integrate the advantages of both random masks and segmentation masks, while avoiding explicit reliance on precise category labels within foreground and background masks, we design and construct Latent Categories Guidance (LCG) to inject only two broad categories—“foreground” and “background”—into a conditioned diffusion model.

Specifically, we categorize masks into four types, as shown in Fig.~\ref{fig:CLCG}, and assign them to either the foreground or background embedding. 
Object semantic masks are assigned to the foreground embedding, as illustrated in Fig.~\ref{fig:CLCG}(a), where benches are masked. The goal is to enable the model to reconstruct foreground objects that are semantically aligned with the masked region \cite{Avrahami2022blenddiff} , utilizing contextual background information.
Scene semantic masks are assigned to the background embedding to enhance the model's ability to reconstruct background semantics.
Similar to previous works \cite{Rombach2022LDM, podell2023sdxl, xie2023smartbrush, wang2023imagen,zhuang2023powerpaint}, we also incorporate random brush masks. However, we assign them only to the background embedding, guiding the model to focus on structural information from the surrounding context.
Finally, to prevent the model from overfitting object semantic masks by binding them too rigidly to foreground objects and ignoring embedding conditions, we assign random object masks to the background embedding as a regularization constraint.
Overall, for the foreground mask $\mathbf{\mathcal{M}}_{fg}$, we directly adopt the object semantic mask, i.e., $\mathbf{\mathcal{M}}_{fg} = \mathbf{\mathcal{M}}_{obj}$. For the background mask, we start with the scene semantic mask $\mathbf{\mathcal{M}}{scene}$ and further incorporate random brush mask $\mathbf{\mathcal{M}}{rand}$ and random object mask $\mathbf{\mathcal{M}}{obj}^\prime$ with probabilities $\mathbf{\mathcal{P}}{rand}$ and $\mathbf{\mathcal{P}}_{obj}$, respectively, as follows:
\begin{equation}
\mathbf{\mathcal{M}}_{bg} = \mathbf{\mathcal{M}}_{scene} + \mathbf{\mathcal{M}}_{rand} \mathbf{\mathcal{P}}_{rand} + \mathbf{\mathcal{M}}_{obj}^\prime \mathbf{\mathcal{P}}_{obj}.
\end{equation}

With this mask assignment strategy, our model learns to capture foreground object distributions under the “foreground” embedding condition, and background semantics and contextual structures under the “background” embedding condition.
During training, we do not explicitly provide category labels as textual prompts to guide generation. Instead, the model learns foreground semantics, background semantics, and contextual structures through learnable embedding weights, ultimately enabling the distribution of both categories to be injected into a single model.
Despite adopting a different training paradigm, our approach remains compatible with other LDM inference schemes \cite{Rombach2022LDM}. By applying Classifier-Free Guidance (CFG) \cite{Ho2021CFG}, we can effectively merge the learned embeddings of the two categories to generate coherent outputs.

\subsection{Structure \& Semantic Consistency Interaction}
\label{sec:interaction}

As shown in Fig.~\ref{fig:pipeline}, we perform multiple interactions between LCG embeddings and latent features during the denoising process. The detailed process of single interaction is illustrated in Fig.~\ref{fig:interaction}, where $\mathbf{L_{in}}$ and $\mathbf{L_{out}}$ represent the input and output features, respectively. The terms $\mathbf{G}_{t}$, $\mathbf{K}_{t}$, $\mathbf{V}_{t}$, $\mathbf{Q}_{t}$, $\mathbf{R}_{t}$, and $\mathbf{S}_{t}$ are all involved in computing the gated linear attention (GLA) \cite{yang2024gla}, where $t$ denotes the index of token.
We first compute self-attention on the normalized $\mathbf{L_{in}}$ using GLA to obtain self-decoded features. Then, following the standard transformer block architecture \cite{dosovitskiy2020vit}, we apply residual connections, normalization, cross-attention, and MLP to produce the final output features $\mathbf{L_{out}}$. Here, LCG embeddings are introduced via cross-attention, enabling the cross-decoding of self-decoded features with embeddings. Throughout the multiple interactions in the pipeline, self-decoding and cross-decoding alternate, and once the embeddings are first introduced via cross-decoding, all subsequent decoding steps incorporate LCG guidance.

Here, we elaborate on the process of computing the self-decoded features $\mathbf{\hat{L}}$ using GLA. We denote the linear projection weights and bias for a variable $x$ as $\mathbf{W}_{x}$ and $\mathbf{b}_{x}$, respectively. First, we compute the query ($\mathbf{Q}$), key ($\mathbf{K}$), and value ($\mathbf{V}$) based on $\mathbf{L_{in}} \in \mathbb{R}^{L \times d}$ (where $L$ is the sequence length and $d$ is the feature dimension) as follows:
\begin{equation}
\begin{aligned}
& \mathbf{Q}=\mathbf{L_{in}} \mathbf{W}_Q \in \mathbb{R}^{L \times d_k}, \\
& \mathbf{K}=\mathbf{L_{in}} \mathbf{W}_K \in \mathbb{R}^{L \times d_k}, \\
& \mathbf{V}=\mathbf{L_{in}} \mathbf{W}_V \in \mathbb{R}^{L \times d_v}, \\
\end{aligned}
\end{equation}
here, ${d_k}$ represents the dimension of $\mathbf{Q}$ and $\mathbf{K}$, while ${d_v}$ represents that of $\mathbf{V}$. The subsequent computation is token-wise, and $t$ is the index of token. We use the 2D forget-gating matrix $\mathbf{G}_t$ to regulate the update of hidden states $\mathbf{S}_{t}$, enabling contextual interactions. The gating matrix $\mathbf{G}_t$ is computed as follows:
\begin{equation}
\begin{aligned}
\mathbf{G}_t & =\bm{\alpha}_t^{\top} \bm{\beta}_t \in \mathbb{R}^{d_k \times d_v}, \\
\bm{\alpha}_t & =\sigma\left(\mathbf{L_{in}} \mathbf{W}_\alpha+\mathbf{b}_\alpha\right)^{\frac{1}{\tau}} \in \mathbb{R}^{L \times d_k}, \\
\bm{\beta}_t & =\sigma\left(\mathbf{L_{in}} \mathbf{W}_\beta+\mathbf{b}_\beta\right)^{\frac{1}{\tau}} \in \mathbb{R}^{L \times d_v},
\end{aligned}
\end{equation}
where $\sigma$ represents the sigmoid function, $\beta$ is the bias term, and $\tau$ is the temperature term. Using $\mathbf{G}_t$ to update the hidden states $\mathbf{S}_{t}$, we obtain $\mathbf{\hat{L}_t}$ as follows:
\begin{equation}
\begin{aligned}
\mathbf{\hat{L}_t} & =\left(\mathbf{R}_t \odot \mathcal{L}\left(\mathbf{O}_t\right)\right) \mathbf{W}_O \in \mathbb{R}^{1 \times d}, \\
\mathbf{R}_t & =\mathcal{S}\left(\mathbf{X}_t \mathbf{W}_r+\mathbf{b}_r\right) \in \mathbb{R}^{1 \times d_v}, \\
\mathbf{O}_t & =\mathbf{Q}_t^{\top} \mathbf{S}_t \in \mathbb{R}^{1 \times d_v}, \\
\mathbf{S}_t & =\mathbf{G}_t \odot \mathbf{S}_{t-1}+\mathbf{K}_t^{\top} V_t \in \mathbb{R}^{d_k \times d_v},
\end{aligned}
\end{equation}
where $\odot$ denotes element-wise multiplication, $\mathcal{S}$ represents the Swish activation function \cite{ramachandran2017swish}, and $\mathcal{L}$ denotes LayerNorm. After computing all $\mathbf{\hat{L}_t}$, we obtain the final self-decoded features $\mathbf{\hat{L}}$.

\section{Experiments}
\label{sec:exs}

\subsection{Datasets and Evaluation}

\paragraph{Construction of Our LCG Training Dataset.}

We define 116 foreground categories and 21 background categories (detailed in the supplementary materials) and adopt an Auto-Labeling framework \cite{sun2023alphaclip} by integrating AlphaCLIP and SAM \cite{Kirillov2023SAM} to obtain fine-grained segmentation masks for foreground and background across multiple datasets.
Specifically, we utilize the following datasets: COCONut-Large \cite{Deng2024coconut} (0.36M images), Object365V2 \cite{shao2019o365} (2.02M images), GoogleLandmarkV2 \cite{weyand2020GLDv2} (4.13M images), and a natural scene dataset (7.49M images) that we curated and collected.
In total, our dataset comprises 14 million images. Following the masking strategy described in Sec.~\ref{sec:CLCG}, we construct a large-scale training dataset aligned with the LCG paradigm, consisting of 4.3M “foreground” image-mask pairs and 9.7M “background” image-mask pairs.

\paragraph{Finetune and Evaluation.}
\label{sec:evaluation}

Recent SOTA inpainting methods \cite{zhuang2023powerpaint, li2022mat, Sargsyan2023migan, suvorov2021lama} are typically evaluated on one or more public benchmarks, including Places2 \cite{zhou2017places}, CelebA-HQ \cite{karras2018celebahq}, and FFHQ \cite{karras2018stylegan_ffhq}. To ensure a comprehensive comparison, we fine-tune PixelHacker on Places2, CelebA-HQ, and FFHQ while strictly following the evaluation protocols used in previous works.
For Places2 (a natural scene dataset), we follow multiple evaluation settings:
(1) Zhuang et al. \cite{zhuang2023powerpaint} and Rombach et al. \cite{Rombach2022LDM} sample 10K test images, apply center cropping to 512×512 resolution, and evaluate performance under 40-50\% masked areas.
(2) Li et al. \cite{li2022mat} use 36.5K validation images, crop them to 512×512, and define two mask configurations (small and large masks) for evaluation.
(3) Sargsyan et al. \cite{Sargsyan2023migan} resize the Places2 validation set to 256×256 and 512×512, applying highly randomized masking strategies for evaluation.
To ensure fair comparisons, we strictly adhere to these evaluation protocols.  
For CelebA-HQ (a human face dataset), we follow Li et al. \cite{li2022mat} and evaluate at 512×512 resolution.
For FFHQ (a human face dataset), we sample 10K images as the evaluation set and follow the masking strategy of Suvorov et al. \cite{suvorov2021lama}, evaluating at 256×256 resolution for comparison. 

\subsection{Implementation Details}

For training on our 14M-image dataset, we use 12 NVIDIA L40S GPUs with a batch size of 528 and a resolution of 512×512. The probabilities $\mathbf{\mathcal{P}}_{rand}$ and $\mathbf{\mathcal{P}}_{obj}$ are both set to 0.5, and the model is trained for 200K iterations.
For fine-tuning on Places2, we use the 1.8M training set of Places2 and fine-tune the model for 120K iterations on 12 NVIDIA L40S GPUs with a batch size 528.
For fine-tuning on CelebA-HQ, we follow the same training set split as Li et al. \cite{li2022mat} and fine-tune the model for 90K iterations on 8 NVIDIA L40S GPUs with a batch size 352.
For fine-tuning on FFHQ, we use 60K images as the training set and 10K images as the evaluation set. The model is fine-tuned for 60K iterations on 12 NVIDIA L40S GPUs with a batch size 528.
In all experiments, we use the pretrained SDXL-VAE \cite{podell2023sdxl} and freeze its parameters. The input resolution is 512×512, the learning rate is set to 1e-5, and we adopt the AdamW optimizer \cite{loshchilov2018AdamW} with betas (0.9, 0.999).

\subsection{Comparison on Places2}

\begin{table}[b]
\centering
\resizebox{0.4\textwidth}{!}{
\begin{tabular}{l|cc}
\toprule
\multirow{2}{*}{\textbf{Method}} & \multicolumn{2}{c}{\textbf{Places2 (Test)}} \\
  &       $\mathbf{FID}\downarrow$    &     $\mathbf{LPIPS}\downarrow$      \\
\midrule
LaMa \cite{suvorov2021lama}                    & 21.07       & 0.2133 \\
LDM  \cite{Rombach2022LDM}                     & 21.42       & 0.2317 \\
SD \cite{Rombach2022LDM}                       & 19.73       & 0.2322 \\
SD(“background”) \cite{Rombach2022LDM}         & 19.21       & 0.2290 \\
SD(“scenery”) \cite{Rombach2022LDM}            & 18.93       & 0.2312 \\
SmartBrush(“scenery”) \cite{xie2023smartbrush} & 87.21       & 0.2812 \\
MI-GAN \cite{Sargsyan2023migan}                & 14.36       & 0.2390 \\
SDXL \cite{podell2023sdxl}                     & \blue{8.66} & 0.2746 \\
PowerPaint \cite{zhuang2023powerpaint}         & 17.91       & 0.2225 \\
PixelHacker(Ours)$^\ddag$                      & 12.19       & \blue{0.2100} \\
PixelHacker(Ours)                              & \red{8.59}  & \red{0.2026} \\
\bottomrule
\end{tabular}
}
\caption{
\textbf{Quantitative comparison of PixelHacker with SOTA methods on 10k samples from Places2 \cite{zhou2017places} test set of 512 resolution with 40-50\%masks.}
Results of other methods except MI-GAN and SDXL are referred from \cite{zhuang2023powerpaint}.
“$\ddag$”: our model without finetune on Places2 \cite{zhou2017places}. 
The best results are in \red{red} while the second best ones are in \blue{blue}.
}
\label{tab:places_4050_mask}
\end{table}

\begin{table*}[t]
\centering
\resizebox{0.95\textwidth}{!}{
\begin{tabular}{l|cccc|cccc}
\toprule
\multirow{2}{*}{\textbf{Method}} &
\multicolumn{4}{c|}{\textbf{Places2 (Large Mask)}} & \multicolumn{4}{c}{\textbf{Places2 (Small Mask)}} \\
& $\mathbf{FID}\downarrow$ & $\mathbf{LPIPS}\downarrow$ & 
        $\mathbf{P}$-$\mathbf{IDS(\%)}\uparrow$ & $\mathbf{U}$-$\mathbf{IDS(\%)}\uparrow$ 
& $\mathbf{FID}\downarrow$ & $\mathbf{LPIPS}\downarrow$ & 
        $\mathbf{P}$-$\mathbf{IDS(\%)}\uparrow$ & $\mathbf{U}$-$\mathbf{IDS(\%)}\uparrow$  \\
\midrule
CoModGAN \cite{zhao2021comodgan}$^\dag$  &  2.92   & 0.192    &\red{19.64} &\blue{35.78} & 1.10 & 0.101 &\blue{26.95}& 41.88 \\
LaMa-Big \cite{suvorov2021lama}$^\dag$   &  2.97   &\red{0.166}& 13.09 & 32.29 &\blue{0.99}&\red{0.086}& 22.79 & 40.58 \\
MADF \cite{zhu2021MADF}                  &  7.53   & 0.181    &  6.00 & 23.78 & 2.24 & 0.095 & 14.85 & 35.03 \\
AOT-GAN \cite{zeng2023AOTGAN}            & 10.64   & 0.195    &  3.07 & 19.92 & 3.19 & 0.101 & 8.07  & 30.94 \\
HiFill \cite{yi2020hifill}               & 28.92   & 0.284    &  1.24 & 11.24 & 7.94 & 0.148 & 3.98  & 23.60 \\
DeepFill v2 \cite{yu2019deepfillv2}      &  9.27   & 0.213    &  4.01 & 21.32 & 3.02 & 0.113 & 9.17  & 32.56 \\
EdgeConnect \cite{nazeri2019edgeconnect} & 12.66   & 0.275    &  1.93 & 15.87 & 4.03 & 0.114 & 5.88  & 27.56 \\
MAT \cite{li2022mat}                     &\blue{2.90}& 0.189 & \blue{19.03} & 35.36 & 1.07 & 0.099 &\red{27.42}&\blue{41.93}\\
PixelHacker(Ours)$^\ddag$           &  3.01   & 0.176 & 13.79 & 33.53 & 1.05 & 0.093 & 23.19 & 40.87 \\
PixelHacker(Ours)                   &\red{2.05}&\blue{0.169}& 16.73 & \red{36.07} & \red{0.82} & \blue{0.088} & 24.78 &\red{42.21} \\ 
\bottomrule
\end{tabular}
}
\caption{
\textbf{Quantitative comparison of PixelHacker with SOTA methods on Places2 \cite{zhou2017places} validation set of 512 resolution with large and small masks follow \cite{li2022mat}.}
Results other than our method referred from \cite{li2022mat}.
“$\dag$”: CoModGAN \cite{zhao2021comodgan} use 8M training images on Places2, LaMa \cite{suvorov2021lama} use 4.5M training images, while other models (without “$\dag$”) including our PixelHacker are trained on 1.8M training images from the standard part of Places2 \cite{zhou2017places}. 
\red{Red}, \blue{blue} and “$\ddag$” are same to  Tab.~\ref{tab:places_4050_mask}.
}
\label{tab:Places_large_small_masks}
\end{table*}

\begin{figure*}
\centering
\includegraphics[width=1.0\linewidth]{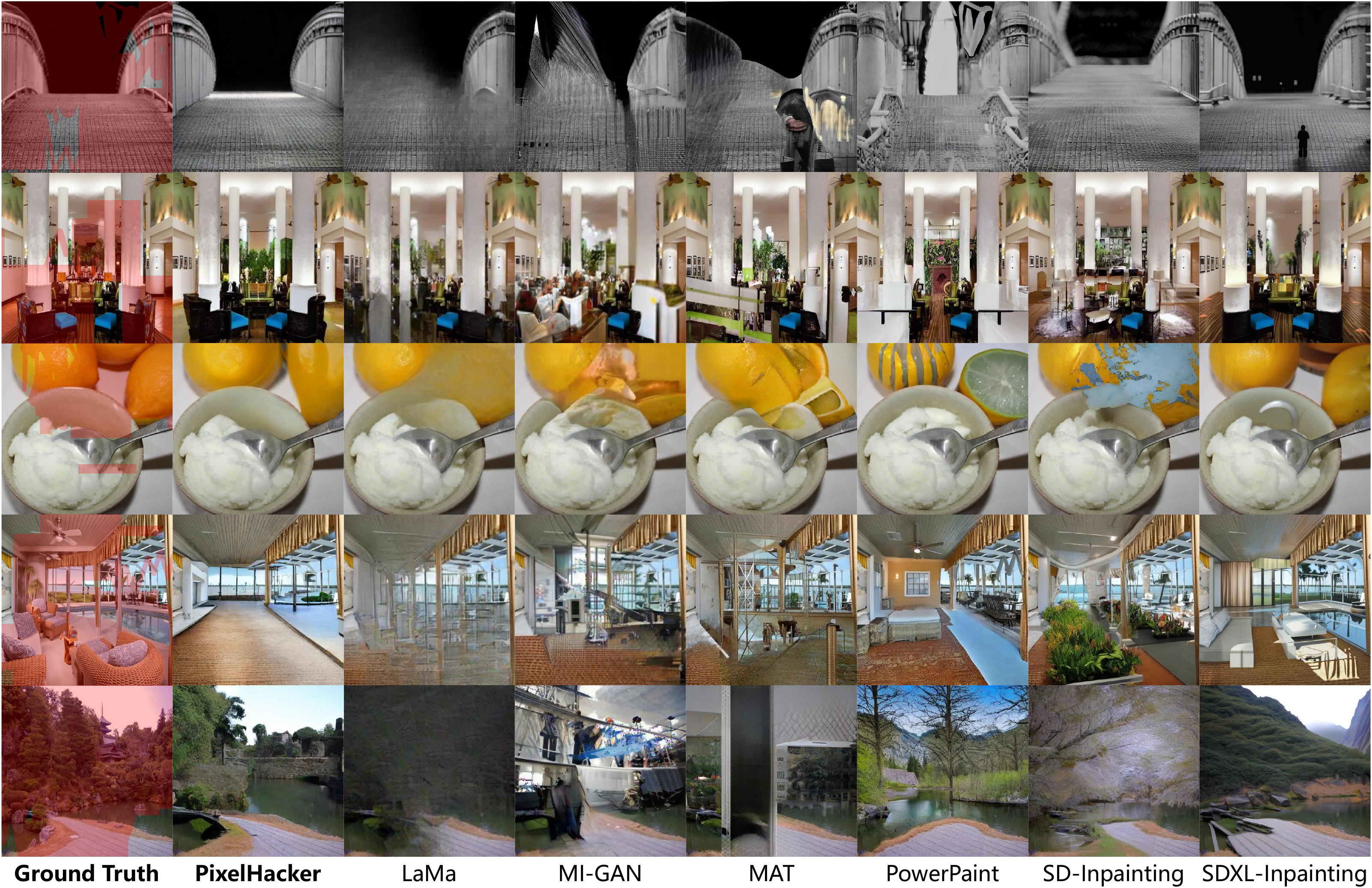}
\vspace{-6mm}
\caption{
\textbf{Qualitative comparison of PixelHacker with SOTA methods on Places2.}
Even under masks that cover almost the entire image, PixelHacker's generated results still exhibit remarkable structural and semantic consistency.
}
\vspace{-3mm}
\label{fig:places_cases}
\end{figure*}

\begin{figure*}[t]
\centering
\includegraphics[width=1.0\linewidth]{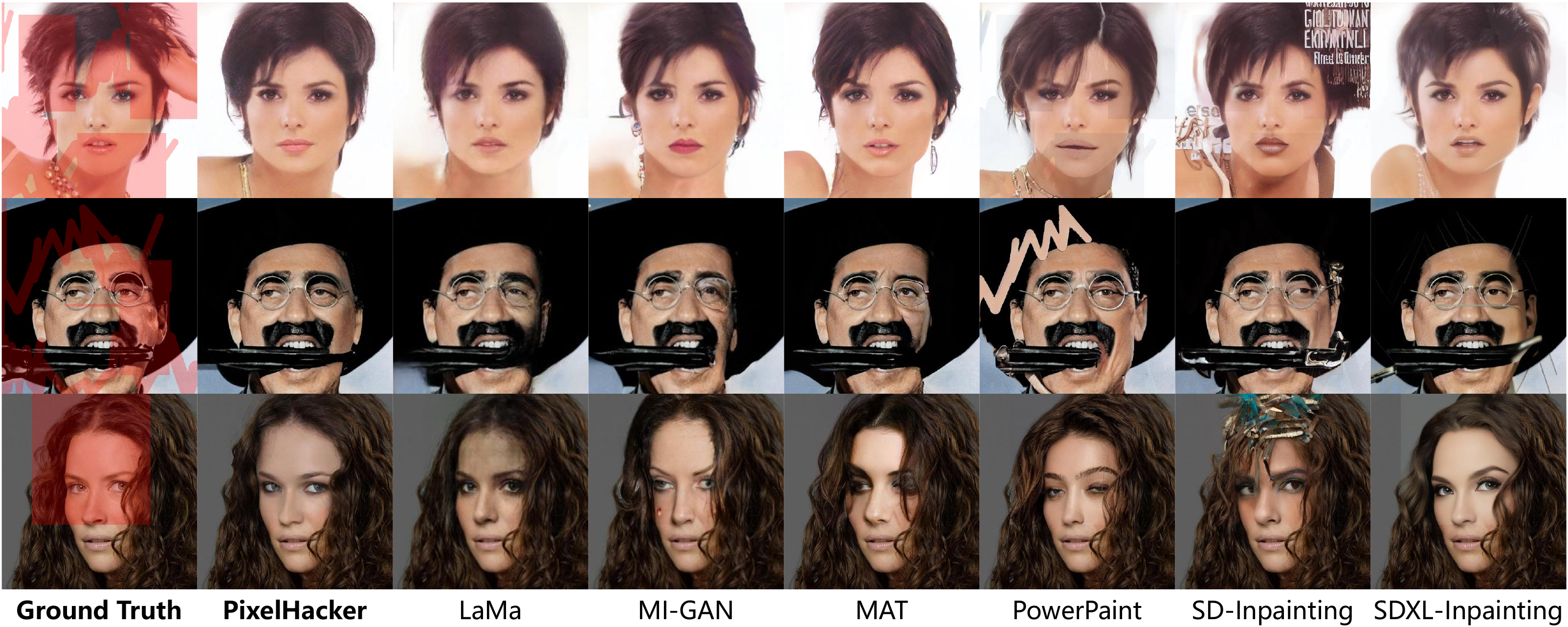}
\vspace{-6mm}
\caption{
\textbf{Qualitative comparison with SOTA methods on CelebA-HQ.}
PixelHacker demonstrates exceptional robustness to mask shapes while maintaining highly consistent semantics, avoiding color discrepancies and artifacts commonly observed in other models.
}
\label{fig:celebahq_cases}
\end{figure*}

We faithfully follow the three evaluation settings on Places2 and conduct fair comparisons with various SOTA methods. The results in Tab.\ref{tab:places_4050_mask}, Tab.\ref{tab:Places_large_small_masks}, and Tab.~\ref{tab:places_migan_mask} correspond to evaluation settings (1), (2), and (3) described in Sec.~\ref{sec:evaluation}, respectively.  
In particular, LDM \cite{Rombach2022LDM} refers to a latent diffusion model fine-tuned for inpainting without text prompts. SD \cite{Rombach2022LDM} denotes SDv1.5-Inpainting, a fine-tuned Stable Diffusion model trained with random masks and image captions for inpainting tasks. SDXL \cite{podell2023sdxl} represents SDXL-Inpainting, a latent text-to-image diffusion model fine-tuned from SDXL-Base \cite{podell2023sdxl} to enable mask-guided inpainting.

First, as shown in Tab.~\ref{tab:places_4050_mask}, we conduct quantitative comparisons on the Places2 test set using 512 resolution with 40-50\% masked areas. The results for MI-GAN \cite{Sargsyan2023migan} and SDXL \cite{podell2023sdxl} are obtained using the official inference codes, while results for other models are taken from \cite{zhuang2023powerpaint}.
Our PixelHacker achieves the best performance, with FID \textbf{8.59} and LPIPS \textbf{0.2026}, surpassing the powerful SD \cite{Rombach2022LDM} and SDXL \cite{podell2023sdxl}. Notably, even without fine-tuning, the zero-shot version of PixelHacker achieves the best LPIPS and second-best FID, trailing only SDXL. This strongly demonstrates the remarkable potential of our paradigm.

\begin{table}[t]
\centering
\resizebox{0.48\textwidth}{!}{
\begin{tabular}{l|cc|cc}
\toprule
\multirow{2}{*}{\textbf{Method}} & 
\multicolumn{2}{c|}{\textbf{Places2 (256)}} & \multicolumn{2}{c}{\textbf{Places2 (512)}} \\
  & $\mathbf{FID}\downarrow$ & $\mathbf{LPIPS}\downarrow$ &
    $\mathbf{FID}\downarrow$ & $\mathbf{LPIPS}\downarrow$ \\
\midrule
LaMa \cite{suvorov2021lama}$^\dag$       & 22.00    & 0.378     & 12.36     & 0.314 \\
CoModGAN \cite{zhao2021comodgan}$^\dag$  &  9.32    & 0.397     &  8.05     & 0.343 \\
SH-GAN \cite{xu2023shgan}                & \red{7.40}& 0.392     &\blue{7.03} & 0.339 \\
ZITS \cite{dong2022zits}                 & 16.78    & \red{0.356} & 12.94    & \blue{0.310} \\
MAT \cite{li2022mat}                     & 14.38    & 0.394     &  8.67      & 0.339 \\
LDM \cite{Rombach2022LDM}                & 13.40    & 0.385     &  8.46      & 0.342 \\
HiFill \cite{yi2020hifill}               & 81.27    & 0.488     & 64.07      & 0.438 \\
MI-GAN \cite{Sargsyan2023migan}          & 11.83    & 0.394     & 10.00      & 0.345 \\
PixelHacker(Ours)$^\ddag$                & 14.60    & 0.380     &  9.40      & 0.317 \\
PixelHacker(Ours)  & \blue{9.25} & \blue{0.367} & \red{5.75} & \red{0.305} \\
\bottomrule
\end{tabular}
}
\caption{
\textbf{Quantitative comparison of PixelHacker with SOTA approaches on Places2 \cite{zhou2017places} validation set of 256 and 512 resolutions follow \cite{Sargsyan2023migan}.}
Results of other methods are referred from \cite{Sargsyan2023migan}.
The images are resized from the High-Resolution version of Places2 dataset\cite{zhou2017places}, into 256 and 512 respectively. 
\red{Red}, \blue{blue} and “$\ddag$” are same to  Tab.~\ref{tab:places_4050_mask}.
“$\dag$” is same to Tab.~\ref{tab:Places_large_small_masks}
}
\vspace{-1mm}
\label{tab:places_migan_mask}
\end{table}

Next, as shown in Tab.~\ref{tab:Places_large_small_masks}, we perform quantitative comparisons on the Places2 validation set, following the large and small mask settings of \cite{li2022mat}, using 512 resolution.
Our PixelHacker, finetuned with only 1.8M images from Places2, achieves the best FID and U-IDS, the second-best LPIPS, and the third-best P-IDS. It is worth noting that: LaMa-Big \cite{suvorov2021lama}, which achieves the best LPIPS, was trained with 4.5M images from Places2. CoModGAN \cite{zhao2021comodgan}, which surpasses PixelHacker in P-IDS, was trained with the full 8M Places2 dataset. Both models utilize significantly more training images from Places2 than PixelHacker, highlighting the data efficiency of our paradigm.

Finally, as shown in Tab.~\ref{tab:places_migan_mask}, we follow \cite{Sargsyan2023migan} and conduct quantitative comparisons on the Places2 validation set at 256 and 512 resolutions.
Our PixelHacker achieves SOTA performance at 512 resolution and second-best results at 256 resolution. Notably, PixelHacker is trained and fine-tuned exclusively at 512 resolution, SH-GAN \cite{xu2023shgan} is trained at both 256 and 512 resolutions, ZITS \cite{dong2022zits} undergoes extensive training across multiple resolutions between 256 and 512. Despite this, PixelHacker still achieves competitive results, demonstrating its robust generalization capability.

We present qualitative comparison results on Places2, as shown in Fig.~\ref{fig:places_cases}. Even under masks that cover almost the entire image, PixelHacker’s generated results still exhibit remarkable structural and semantic consistency. Unlike LaMa \cite{suvorov2021lama} and MI-GAN \cite{Sargsyan2023migan}, which produce artifacts and ambiguous generations (e.g., the first row), PixelHacker does not suffer from such issues. Under medium-sized masks, even in semantically rich and structurally complex scenes, PixelHacker consistently maintains structural coherence (e.g., the second and third rows).

\subsection{Comparison on CelebA-HQ and FFHQ}
\label{sec:CelebA-HQ and FFHQ}

\textbf{CelebA-HQ.} As shown in Tab.~\ref{tab:Celebahq_512_masks}, we follow \cite{li2022mat} and conduct quantitative comparisons on CelebA-HQ using 512 resolution. PixelHacker consistently achieves SOTA performance, demonstrating its strong generalization ability when transferring from natural scenes to facial image domains under our LCG paradigm.
We further provide qualitative comparisons, as shown in Fig.~\ref{fig:celebahq_cases}. PixelHacker generates clear and well-formed facial features without noticeable distortions, maintaining strong semantic consistency. Moreover, PixelHacker does not introduce abrupt or irrelevant textures, indicating superior structural consistency.

\begin{figure}[t]
\centering
\includegraphics[width=1.0\linewidth]{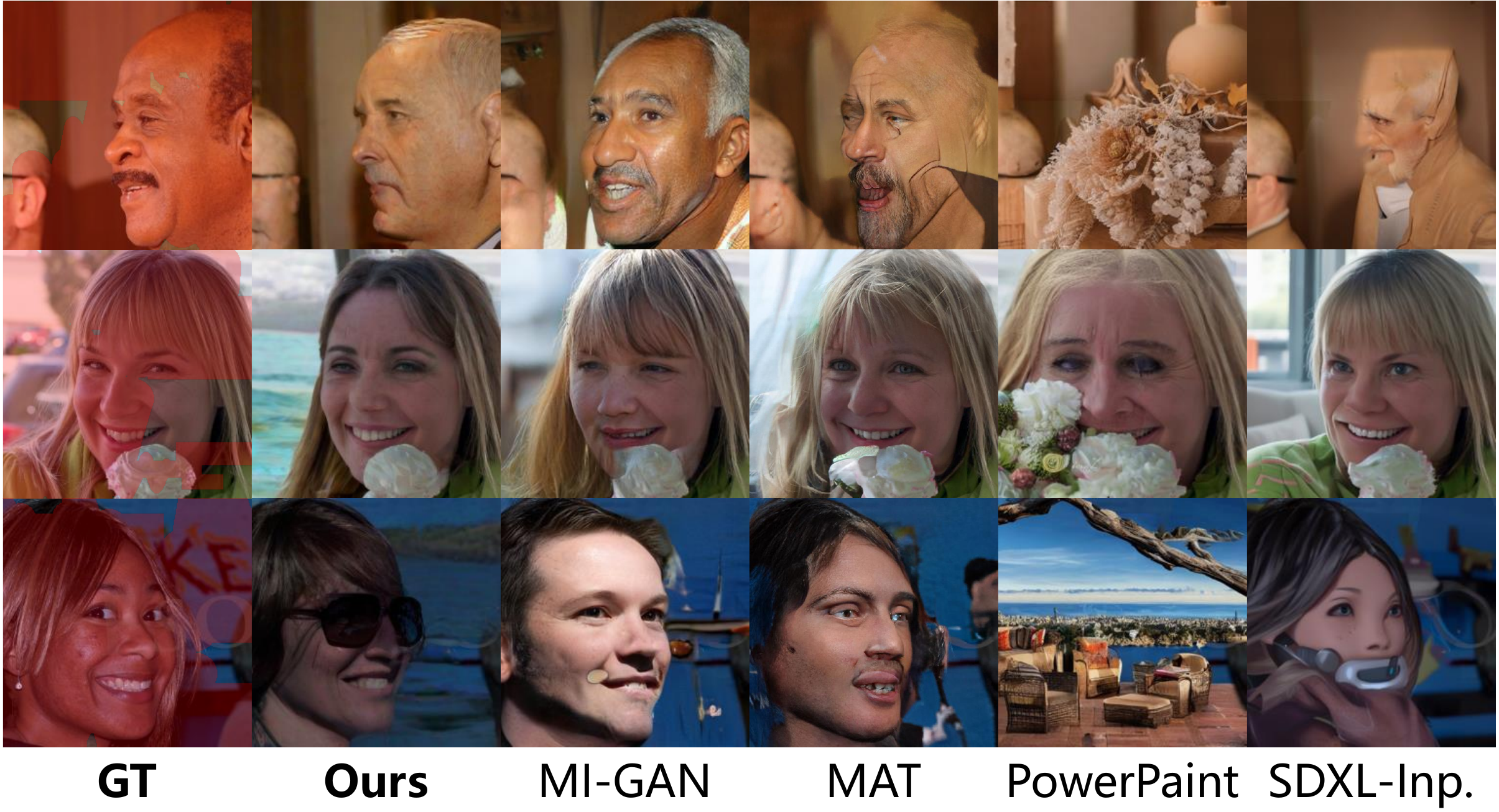}
\vspace{-6mm}
\caption{\textbf{Qualitative comparison with SOTA methods on FFHQ.}
Our PixelHacker generates more realistic results compared to other methods and exhibits strong adaptability to scenes with complex hierarchies and challenging lighting conditions.
}
\label{fig:ffhq_cases}
\end{figure}

\begin{table}
\centering
\resizebox{0.31\textwidth}{!}{
    \begin{tabular}{l|cc}
    \toprule
    \multirow{2}{*}{\textbf{Method}} & \multicolumn{2}{c}{\textbf{CelebA-HQ}} \\
     & $\mathbf{FID}\downarrow$ & $\mathbf{LPIPS}\downarrow$  \\
    \midrule
    CoModGAN \cite{zhao2021comodgan}         &  5.65 & 0.140 \\
    LaMa \cite{suvorov2021lama}              &  8.15 & 0.143 \\
    ICT  \cite{wan2021ICT}                   & 12.84 & 0.195 \\
    MADF \cite{zhu2021MADF}                  &  6.83 & 0.130 \\
    AOT-GAN \cite{zeng2023AOTGAN}            & 10.82 & 0.145 \\
    DeepFill v2 \cite{yu2019deepfillv2}      & 24.42 & 0.221 \\
    EdgeConnect \cite{nazeri2019edgeconnect} & 39.99 & 0.208 \\
    MAT \cite{li2022mat}                     &  4.86 & 0.125 \\
    PixelHacker(Ours)                        & \red{4.75} & \red{0.115}   \\ 
    \bottomrule
    \end{tabular}
}
\caption{\textbf{Quantitative comparison with SOTA methods on CelebA-HQ in 512 resolution follow \cite{li2022mat}.}
Results of other methods are referred from \cite{li2022mat}. The best results are in \red{red}.
}
\label{tab:Celebahq_512_masks}
\end{table}

\begin{table}[t]
\centering
\resizebox{0.31\textwidth}{!}{
    \begin{tabular}{l|cc}
    \toprule
    \multirow{2}{*}{\textbf{Method}} & \multicolumn{2}{c}{\textbf{FFHQ}} \\
     &  $\mathbf{FID}\downarrow$ & $\mathbf{LPIPS}\downarrow$ \\
    \midrule
    LaMa \cite{suvorov2021lama}            & 52.17 & 0.366 \\
    MI-GAN \cite{Sargsyan2023migan}        & 27.65 & 0.358 \\
    SD \cite{Rombach2022LDM}               & 40.24 & 0.359 \\
    PowerPaint \cite{zhuang2023powerpaint} & 38.25 & 0.409 \\
    SDXL \cite{podell2023sdxl}             & 12.36 & 0.278 \\
    PixelHacker(Ours)                      & \red{6.35} & \red{0.229} \\
    \bottomrule
    \end{tabular}
}
\caption{
\textbf{Quantitative comparison with SOTA methods on FFHQ in 256 resolution.} 
The best results are in \red{red}.
}
\vspace{-3mm}
\label{tab:ffhq}
\end{table}

\textbf{FFHQ.} Tab.~\ref{tab:ffhq} presents PixelHacker’s SOTA quantitative results on FFHQ \cite{karras2018stylegan_ffhq}, demonstrating its remarkable generalization ability to lower resolutions despite being exclusively trained at 512 resolution.
Fig.~\ref{fig:ffhq_cases} provides qualitative comparisons, illustrating that PixelHacker produces more realistic results than other methods while exhibiting strong adaptability to complex scene hierarchies and challenging lighting conditions.

\subsection{Ablation Study}

\textbf{Ablation of various masks in LCG.}
We evaluate the impact of the mask construction strategy proposed in Sec.~\ref{sec:CLCG}, as shown in Tab.~\ref{tab:ablation_CLCG}. For each setting, we fairly initialize the model using weights trained for 90K iterations on our 14M dataset, then continue training 12K iterations before evaluating on a 3K-image subset of the Places2 validation set \cite{zhou2017places}. The results demonstrate that using $\mathcal{M}_{obj}$, $\mathcal{M}_{scene}$, $\mathcal{M}_{rand}$, and $\mathcal{M}_{obj}^\prime$ yields the best performance. 

\begin{table}[t]
\centering
\resizebox{0.40\textwidth}{!}{
\begin{tabular}{cccc|cc}
\toprule
   $\mathcal{M}_{obj}$ & 
   $\mathcal{M}_{scene}$ & $\mathcal{M}_{rand}$ & $\mathcal{M}_{obj}^\prime$ & 
   $\mathbf{FID}\downarrow$ & $\mathbf{LPIPS}\downarrow$ \\
\midrule
\checkmark & \checkmark &            &            & 12.69       &      0.1758  \\
\checkmark & \checkmark & \checkmark &            & 12.65       & \red{0.1735} \\
\checkmark & \checkmark & \checkmark & \checkmark & \red{12.62} & \red{0.1735} \\
\bottomrule
\end{tabular}
}
\vspace{-1mm}
\caption{
\textbf{Ablation of various masks in LCG.} Best results in \red{red}.  
}
\label{tab:ablation_CLCG}
\end{table}

\textbf{Ablation of embedding dims.}
To investigate whether increasing the embedding dimension (E.Dim) leads to significant performance improvements, we compare various E.Dim values of 20, 64, 256, and 1024, as presented in Tab.~\ref{tab:ablation_embedding_dim}. For each configuration, we fairly initialize the model using weights trained for 200K iterations on our 14M dataset, then continue training 36K iterations, evaluating on custom 10K images.
The results indicate that scaling up E.Dim does not significantly enhance performance. This suggests that a smaller E.Dim is sufficient to represent latent “foreground” and “background” features. Based on these findings, we adopt E.Dim = 20 as the default setting.

\begin{table}[t]
\centering
\resizebox{0.41\textwidth}{!}{
\begin{tabular}{c | cc || c | cc}
    \hline
    \textbf{\textit{E.Dim}}  & $\mathbf{FID}\downarrow$ & $\mathbf{LPIPS}\downarrow$ &
    \textbf{\textit{E.Dim}}  & $\mathbf{FID}\downarrow$ & $\mathbf{LPIPS}\downarrow$ \\
    \hline
    20$^\dag$ & 0.755       & 0.264 & 256  & 0.754  & 0.258 \\
    64        & \red{0.735} & 0.257 & 1024 & 0.744  & \red{0.255} \\
    \hline
\end{tabular}
}
\vspace{-1mm}
\caption{
\textbf{Ablation of embedding dims.} 
“$\dag$”: we use 20 as the default size in PixelHacker. 
The best results are in \red{red}.
}
\label{tab:ablation_embedding_dim}
\end{table}

\textbf{Ablation of guidance scales.}
Similar to other Latent Diffusion Models, we employ Classifier-Free Guidance inference \cite{Ho2021CFG}. Here, to assess the impact of guidance scale on generation quality, we compare multiple scales ranging from 1.0 (no guidance) to 4.0, as shown in Tab.~\ref{tab:ablation_guidance_scale}.
For each configuration, we initialize the model using weights that fine-tune on CelebA-HQ for 21K iterations, and evaluate on the CelebA-HQ validation set used in Sec.~\ref{sec:CelebA-HQ and FFHQ}. The results confirm that 2.0 is the best, which we adopt as default.

\begin{table}[t]
\centering
\resizebox{0.47\textwidth}{!}{
\begin{tabular}{l|cccccccc}
\toprule
$\textbf{\textit{Scale}}$   &  1.0  &   1.5 & 2.0    & 2.5   & 3.0    & 3.5   & 4.0        \\  
\midrule
$\mathbf{FID}  \downarrow$  &  9.70 &  8.60 & \red{8.44} & 8.47  & 8.48   & 8.51  & 8.73 \\ 
$\mathbf{LPIPS}\downarrow$  & 0.124 & 0.119 & \red{0.118}& 0.119 & 0.120  & 0.121 & 0.123 \\  
\bottomrule
\end{tabular}
}
\vspace{-1mm}
\caption{
\textbf{Ablation of guidance scales.} The best scale is 2.0.
}
\vspace{-6mm}
\label{tab:ablation_guidance_scale}  
\end{table}

\section{Conclusion}
\label{sec:conclusion}

In this work, we introduce Latent Categories Guidance (LCG), a simple yet effective inpainting paradigm that guides the model toward structural and semantic consistency through latent foreground and background features. Then, we propose PixelHacker, a diffusion-based inpainting model trained with LCG on 14M image-mask pairs and fine-tuning on open-source benchmarks. Extensive experiments demonstrate that PixelHacker consistently achieves SOTA performance across various benchmarks.

\small
\bibliographystyle{ieeenat_fullname}
\bibliography{main}

\begin{thebibliography}{40}
\providecommand{\natexlab}[1]{#1}
\providecommand{\url}[1]{\texttt{#1}}
\expandafter\ifx\csname urlstyle\endcsname\relax
  \providecommand{\doi}[1]{doi: #1}\else
  \providecommand{\doi}{doi: \begingroup \urlstyle{rm}\Url}\fi

\bibitem[Avrahami et~al.(2022)Avrahami, Lischinski, and Fried]{Avrahami2022blenddiff}
Omri Avrahami, Dani Lischinski, and Ohad Fried.
\newblock Blended diffusion for text-driven editing of natural images.
\newblock In \emph{Proceedings of the IEEE/CVF Conference on Computer Vision and Pattern Recognition (CVPR)}, pages 18208--18218, 2022.

\bibitem[Bertalmio et~al.(2000)Bertalmio, Sapiro, Caselles, and Ballester]{Bertalmio2000imageinpaint}
Marcelo Bertalmio, Guillermo Sapiro, Vincent Caselles, and Coloma Ballester.
\newblock Image inpainting.
\newblock In \emph{Proceedings of the 27th Annual Conference on Computer Graphics and Interactive Techniques}, page 417–424, USA, 2000. ACM Press/Addison-Wesley Publishing Co.

\bibitem[Choi et~al.(2018)Choi, Choi, Kim, Ha, Kim, and Choo]{Choi2018stargan}
Yunjey Choi, Minje Choi, Munyoung Kim, Jung-Woo Ha, Sunghun Kim, and Jaegul Choo.
\newblock Stargan: Unified generative adversarial networks for multi-domain image-to-image translation.
\newblock In \emph{2018 IEEE/CVF Conference on Computer Vision and Pattern Recognition}, pages 8789--8797, 2018.

\bibitem[Choi et~al.(2020)Choi, Uh, Yoo, and Ha]{Choi2020starganv2}
Yunjey Choi, Youngjung Uh, Jaejun Yoo, and Jung-Woo Ha.
\newblock Stargan v2: Diverse image synthesis for multiple domains.
\newblock In \emph{2020 IEEE/CVF Conference on Computer Vision and Pattern Recognition (CVPR)}, pages 8185--8194, 2020.

\bibitem[Deng et~al.(2024)Deng, Yu, Wang, Shen, and Chen]{Deng2024coconut}
Xueqing Deng, Qihang Yu, Peng Wang, Xiaohui Shen, and Liang-Chieh Chen.
\newblock Coconut: Modernizing coco segmentation.
\newblock In \emph{Proceedings of the IEEE/CVF Conference on Computer Vision and Pattern Recognition}, 2024.

\bibitem[Dhariwal and Nichol(2021)]{Dhariwal2021CG}
Prafulla Dhariwal and Alexander Nichol.
\newblock Diffusion models beat gans on image synthesis.
\newblock In \emph{Advances in Neural Information Processing Systems}, pages 8780--8794. Curran Associates, Inc., 2021.

\bibitem[Dong et~al.(2022)Dong, Cao, and Fu]{dong2022zits}
Qiaole Dong, Chenjie Cao, and Yanwei Fu.
\newblock Incremental transformer structure enhanced image inpainting with masking positional encoding.
\newblock In \emph{Proceedings of the IEEE/CVF Conference on Computer Vision and Pattern Recognition}, 2022.

\bibitem[Dosovitskiy et~al.(2021)Dosovitskiy, Beyer, Kolesnikov, Weissenborn, Zhai, Unterthiner, Dehghani, Minderer, Heigold, Gelly, Uszkoreit, and Houlsby]{dosovitskiy2020vit}
Alexey Dosovitskiy, Lucas Beyer, Alexander Kolesnikov, Dirk Weissenborn, Xiaohua Zhai, Thomas Unterthiner, Mostafa Dehghani, Matthias Minderer, Georg Heigold, Sylvain Gelly, Jakob Uszkoreit, and Neil Houlsby.
\newblock An image is worth 16x16 words: Transformers for image recognition at scale.
\newblock In \emph{International Conference on Learning Representations}, 2021.

\bibitem[Ho and Salimans(2021)]{Ho2021CFG}
Jonathan Ho and Tim Salimans.
\newblock Classifier-free diffusion guidance.
\newblock In \emph{NeurIPS 2021 Workshop on Deep Generative Models and Downstream Applications}, 2021.

\bibitem[Karras et~al.(2020)Karras, Laine, Aittala, Hellsten, Lehtinen, and Aila]{Karras2019stylegan2}
Tero Karras, Samuli Laine, Miika Aittala, Janne Hellsten, Jaakko Lehtinen, and Timo Aila.
\newblock Analyzing and improving the image quality of {StyleGAN}.
\newblock In \emph{Proc. CVPR}, 2020.

\bibitem[Karras et~al.(2021)Karras, Laine, and Aila]{karras2018stylegan_ffhq}
Tero Karras, Samuli Laine, and Timo Aila.
\newblock A style-based generator architecture for generative adversarial networks.
\newblock \emph{IEEE Transactions on Pattern Analysis and Machine Intelligence}, 43\penalty0 (12):\penalty0 4217--4228, 2021.

\bibitem[Kingma and Welling(2022)]{kingma2022VAE}
Diederik~P Kingma and Max Welling.
\newblock Auto-encoding variational bayes, 2022.

\bibitem[Kirillov et~al.(2023)Kirillov, Mintun, Ravi, Mao, Rolland, Gustafson, Xiao, Whitehead, Berg, Lo, Dollar, and Girshick]{Kirillov2023SAM}
Alexander Kirillov, Eric Mintun, Nikhila Ravi, Hanzi Mao, Chloe Rolland, Laura Gustafson, Tete Xiao, Spencer Whitehead, Alexander~C. Berg, Wan-Yen Lo, Piotr Dollar, and Ross Girshick.
\newblock Segment anything.
\newblock In \emph{Proceedings of the IEEE/CVF International Conference on Computer Vision (ICCV)}, pages 4015--4026, 2023.

\bibitem[Li et~al.(2022)Li, Lin, Zhou, Qi, Wang, and Jia]{li2022mat}
Wenbo Li, Zhe Lin, Kun Zhou, Lu Qi, Yi Wang, and Jiaya Jia.
\newblock Mat: Mask-aware transformer for large hole image inpainting.
\newblock In \emph{Proceedings of the IEEE/CVF Conference on Computer Vision and Pattern Recognition}, 2022.

\bibitem[Liu et~al.(2018)Liu, Reda, Shih, Wang, Tao, and Catanzaro]{liu2018partialinpainting}
Guilin Liu, Fitsum~A. Reda, Kevin~J. Shih, Ting-Chun Wang, Andrew Tao, and Bryan Catanzaro.
\newblock Image inpainting for irregular holes using partial convolutions.
\newblock In \emph{The European Conference on Computer Vision (ECCV)}, 2018.

\bibitem[Loshchilov and Hutter(2019)]{loshchilov2018AdamW}
Ilya Loshchilov and Frank Hutter.
\newblock Decoupled weight decay regularization.
\newblock In \emph{International Conference on Learning Representations}, 2019.

\bibitem[Nazeri et~al.(2019)Nazeri, Ng, Joseph, Qureshi, and Ebrahimi]{nazeri2019edgeconnect}
Kamyar Nazeri, Eric Ng, Tony Joseph, Faisal Qureshi, and Mehran Ebrahimi.
\newblock Edgeconnect: Structure guided image inpainting using edge prediction.
\newblock In \emph{The IEEE International Conference on Computer Vision (ICCV) Workshops}, 2019.

\bibitem[Pathak et~al.(2016)Pathak, Krähenbühl, Donahue, Darrell, and Efros]{Pathak2016CtxtEnc}
Deepak Pathak, Philipp Krähenbühl, Jeff Donahue, Trevor Darrell, and Alexei~A. Efros.
\newblock Context encoders: Feature learning by inpainting.
\newblock In \emph{2016 IEEE Conference on Computer Vision and Pattern Recognition (CVPR)}, pages 2536--2544, 2016.

\bibitem[Podell et~al.(2023)Podell, English, Lacey, Blattmann, Dockhorn, Müller, Penna, and Rombach]{podell2023sdxl}
Dustin Podell, Zion English, Kyle Lacey, Andreas Blattmann, Tim Dockhorn, Jonas Müller, Joe Penna, and Robin Rombach.
\newblock Sdxl: Improving latent diffusion models for high-resolution image synthesis, 2023.

\bibitem[Ramachandran et~al.(2017)Ramachandran, Zoph, and Le]{ramachandran2017swish}
Prajit Ramachandran, Barret Zoph, and Quoc~V Le.
\newblock Searching for activation functions.
\newblock \emph{arXiv preprint arXiv:1710.05941}, 2017.

\bibitem[Rombach et~al.(2022)Rombach, Blattmann, Lorenz, Esser, and Ommer]{Rombach2022LDM}
Robin Rombach, Andreas Blattmann, Dominik Lorenz, Patrick Esser, and Bj\"orn Ommer.
\newblock High-resolution image synthesis with latent diffusion models.
\newblock In \emph{Proceedings of the IEEE/CVF Conference on Computer Vision and Pattern Recognition (CVPR)}, pages 10684--10695, 2022.

\bibitem[Sargsyan et~al.(2023)Sargsyan, Navasardyan, Xu, and Shi]{Sargsyan2023migan}
Andranik Sargsyan, Shant Navasardyan, Xingqian Xu, and Humphrey Shi.
\newblock Mi-gan: A simple baseline for image inpainting on mobile devices.
\newblock In \emph{Proceedings of the IEEE/CVF International Conference on Computer Vision (ICCV)}, pages 7335--7345, 2023.

\bibitem[Shao et~al.(2019)Shao, Li, Zhang, Peng, Yu, Zhang, Li, and Sun]{shao2019o365}
Shuai Shao, Zeming Li, Tianyuan Zhang, Chao Peng, Gang Yu, Xiangyu Zhang, Jing Li, and Jian Sun.
\newblock Objects365: A large-scale, high-quality dataset for object detection.
\newblock In \emph{2019 IEEE/CVF International Conference on Computer Vision (ICCV)}, pages 8429--8438, 2019.

\bibitem[Sun et~al.(2023)Sun, Fang, Wu, Zhang, Zang, Kong, Xiong, Lin, and Wang]{sun2023alphaclip}
Zeyi Sun, Ye Fang, Tong Wu, Pan Zhang, Yuhang Zang, Shu Kong, Yuanjun Xiong, Dahua Lin, and Jiaqi Wang.
\newblock Alpha-clip: A clip model focusing on wherever you want, 2023.

\bibitem[Suvorov et~al.(2021)Suvorov, Logacheva, Mashikhin, Remizova, Ashukha, Silvestrov, Kong, Goka, Park, and Lempitsky]{suvorov2021lama}
Roman Suvorov, Elizaveta Logacheva, Anton Mashikhin, Anastasia Remizova, Arsenii Ashukha, Aleksei Silvestrov, Naejin Kong, Harshith Goka, Kiwoong Park, and Victor Lempitsky.
\newblock Resolution-robust large mask inpainting with fourier convolutions.
\newblock \emph{arXiv preprint arXiv:2109.07161}, 2021.

\bibitem[Tero~Karras and Lehtinen(2018)]{karras2018celebahq}
Samuli~Laine Tero~Karras, Timo~Aila and Jaakko Lehtinen.
\newblock Progressive growing of gans for improved quality, stability and variation.
\newblock In \emph{International Conference on Learning Representations (ICLR)}, 2018.

\bibitem[Wan et~al.(2021)Wan, Zhang, Chen, and Liao]{wan2021ICT}
Ziyu Wan, Jingbo Zhang, Dongdong Chen, and Jing Liao.
\newblock High-fidelity pluralistic image completion with transformers.
\newblock In \emph{2021 IEEE/CVF International Conference on Computer Vision (ICCV)}, pages 4672--4681, 2021.

\bibitem[Wang et~al.(2023)Wang, Saharia, Montgomery, Pont-Tuset, Noy, Pellegrini, Onoe, Laszlo, Fleet, Soricut, et~al.]{wang2023imagen}
Su Wang, Chitwan Saharia, Ceslee Montgomery, Jordi Pont-Tuset, Shai Noy, Stefano Pellegrini, Yasumasa Onoe, Sarah Laszlo, David~J Fleet, Radu Soricut, et~al.
\newblock Imagen editor and editbench: Advancing and evaluating text-guided image inpainting.
\newblock In \emph{Proceedings of the IEEE/CVF Conference on Computer Vision and Pattern Recognition}, pages 18359--18369, 2023.

\bibitem[Weyand et~al.(2020)Weyand, Araujo, Cao, and Sim]{weyand2020GLDv2}
Tobias Weyand, André Araujo, Bingyi Cao, and Jack Sim.
\newblock Google landmarks dataset v2 – a large-scale benchmark for instance-level recognition and retrieval.
\newblock In \emph{2020 IEEE/CVF Conference on Computer Vision and Pattern Recognition (CVPR)}, pages 2572--2581, 2020.

\bibitem[Xie et~al.(2023)Xie, Zhang, Lin, Hinz, and Zhang]{xie2023smartbrush}
Shaoan Xie, Zhifei Zhang, Zhe Lin, Tobias Hinz, and Kun Zhang.
\newblock Smartbrush: Text and shape guided object inpainting with diffusion model.
\newblock In \emph{Proceedings of the IEEE/CVF Conference on Computer Vision and Pattern Recognition}, pages 22428--22437, 2023.

\bibitem[Xu et~al.(2023)Xu, Navasardyan, Tadevosyan, Sargsyan, Mu, and Shi]{xu2023shgan}
Xingqian Xu, Shant Navasardyan, Vahram Tadevosyan, Andranik Sargsyan, Yadong Mu, and Humphrey Shi.
\newblock Image completion with heterogeneously filtered spectral hints.
\newblock In \emph{Proceedings of the IEEE/CVF Winter Conference on Applications of Computer Vision}, pages 4591--4601, 2023.

\bibitem[Yang et~al.(2024)Yang, Wang, Shen, Panda, and Kim]{yang2024gla}
Songlin Yang, Bailin Wang, Yikang Shen, Rameswar Panda, and Yoon Kim.
\newblock Gated linear attention transformers with hardware-efficient training.
\newblock In \emph{Proceedings of ICML}, 2024.

\bibitem[Yi et~al.(2020)Yi, Tang, Azizi, Jang, and Xu]{yi2020hifill}
Zili Yi, Qiang Tang, Shekoofeh Azizi, Daesik Jang, and Zhan Xu.
\newblock Contextual residual aggregation for ultra high-resolution image inpainting.
\newblock In \emph{Proceedings of the IEEE/CVF Conference on Computer Vision and Pattern Recognition}, pages 7508--7517, 2020.

\bibitem[Yu et~al.(2018)Yu, Lin, Yang, Shen, Lu, and Huang]{yu2018deepfillv1}
Jiahui Yu, Zhe Lin, Jimei Yang, Xiaohui Shen, Xin Lu, and Thomas~S. Huang.
\newblock Generative image inpainting with contextual attention.
\newblock In \emph{2018 IEEE/CVF Conference on Computer Vision and Pattern Recognition}, pages 5505--5514, 2018.

\bibitem[Yu et~al.(2019)Yu, Lin, Yang, Shen, Lu, and Huang]{yu2019deepfillv2}
Jiahui Yu, Zhe Lin, Jimei Yang, Xiaohui Shen, Xin Lu, and Thomas Huang.
\newblock Free-form image inpainting with gated convolution.
\newblock In \emph{2019 IEEE/CVF International Conference on Computer Vision (ICCV)}, pages 4470--4479, 2019.

\bibitem[Zeng et~al.(2023)Zeng, Fu, Chao, and Guo]{zeng2023AOTGAN}
Yanhong Zeng, Jianlong Fu, Hongyang Chao, and Baining Guo.
\newblock Aggregated contextual transformations for high-resolution image inpainting.
\newblock \emph{IEEE Transactions on Visualization and Computer Graphics}, 29\penalty0 (7):\penalty0 3266--3280, 2023.

\bibitem[Zhao et~al.(2021)Zhao, Cui, Sheng, Dong, Liang, Chang, and Xu]{zhao2021comodgan}
Shengyu Zhao, Jonathan Cui, Yilun Sheng, Yue Dong, Xiao Liang, Eric~I Chang, and Yan Xu.
\newblock Large scale image completion via co-modulated generative adversarial networks.
\newblock In \emph{International Conference on Learning Representations (ICLR)}, 2021.

\bibitem[Zhou et~al.(2017)Zhou, Lapedriza, Khosla, Oliva, and Torralba]{zhou2017places}
Bolei Zhou, Agata Lapedriza, Aditya Khosla, Aude Oliva, and Antonio Torralba.
\newblock Places: A 10 million image database for scene recognition.
\newblock \emph{IEEE Transactions on Pattern Analysis and Machine Intelligence}, 2017.

\bibitem[Zhu et~al.(2021)Zhu, He, Li, Li, Li, Liu, Ding, and Zhang]{zhu2021MADF}
Manyu Zhu, Dongliang He, Xin Li, Chao Li, Fu Li, Xiao Liu, Errui Ding, and Zhaoxiang Zhang.
\newblock Image inpainting by end-to-end cascaded refinement with mask awareness.
\newblock \emph{IEEE Transactions on Image Processing}, 30:\penalty0 4855--4866, 2021.

\bibitem[Zhuang et~al.(2023)Zhuang, Zeng, Liu, Yuan, and Chen]{zhuang2023powerpaint}
Junhao Zhuang, Yanhong Zeng, Wenran Liu, Chun Yuan, and Kai Chen.
\newblock A task is worth one word: Learning with task prompts for high-quality versatile image inpainting, 2023.

\end{thebibliography}

\newpage
\begin{figure*}[b]
    \centering
    \includegraphics[width=1.0\linewidth]{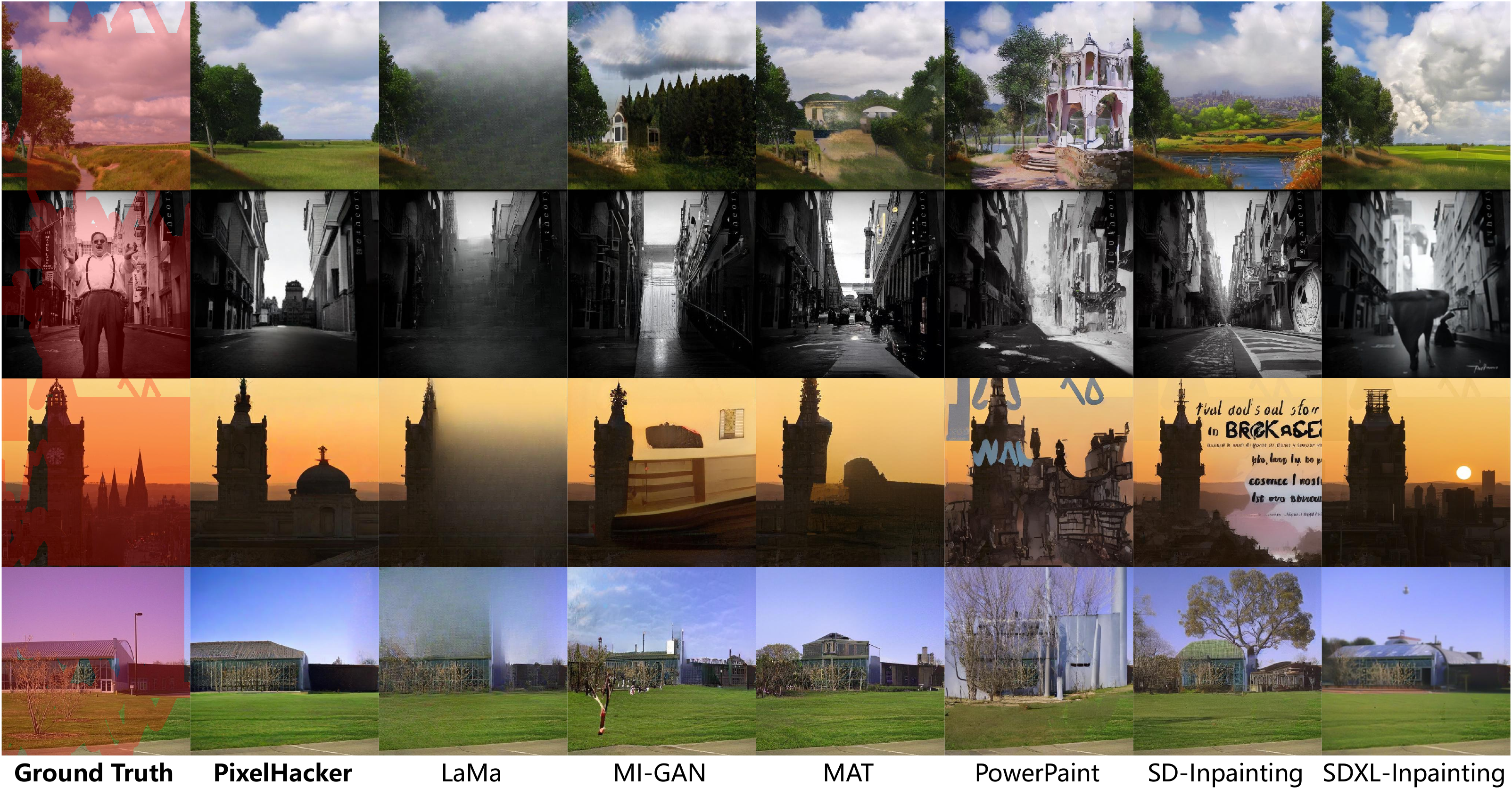}
    \caption{\textbf{More qualitative results of our PixelHacker vs. other SOTA methods \cite{suvorov2021lama,Sargsyan2023migan, li2022mat,zhuang2023powerpaint,Rombach2022LDM, podell2023sdxl} on Places2 \cite{zhou2017places}.} Even under masks that obscure almost the entire image, PixelHacker consistently maintains remarkable semantic and structural coherence.}
    \label{fig:places_sup}
\end{figure*}

\begin{figure*}[b]
\centering
\includegraphics[width=1.0\linewidth]{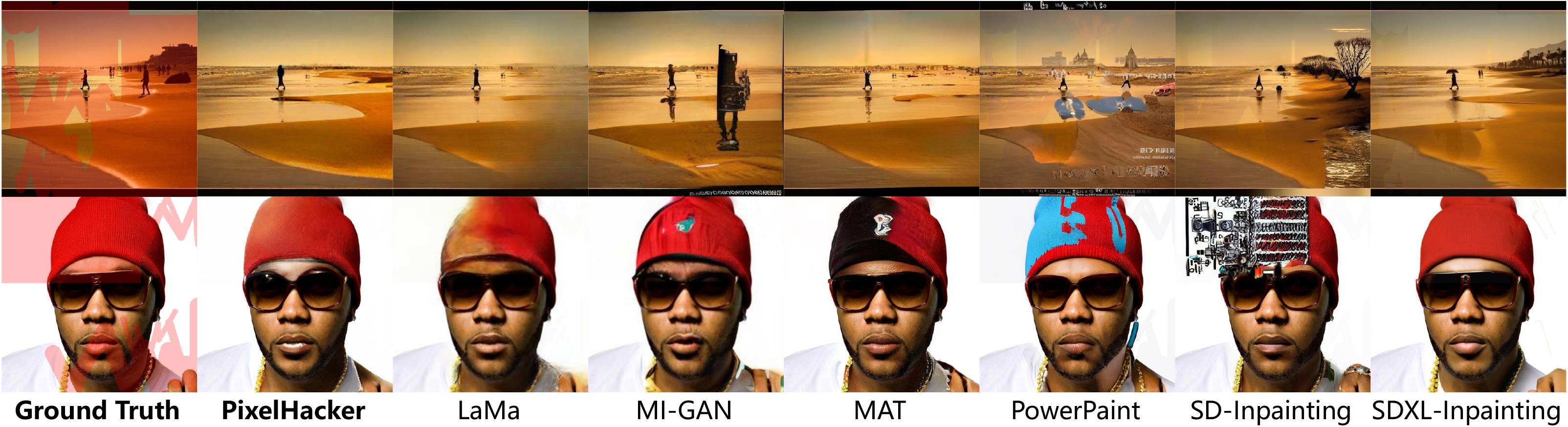}
\caption{\textbf{Failure cases on natural and human-face scenes.} In the first row, our PixelHacker struggles with generating fine details for the small black humanoid silhouette; in the second row, PixelHacker exhibits suboptimal reconstruction of the fingers. Nevertheless, it still maintains significantly better semantic and structural coherence compared to all other SOTA methods.}   
\label{fig:badcase}
\end{figure*}

\section*{Supplementary Materials}
\paragraph{Potential Categories.}
(1) “foreground”:
person, bicycle, car, motorcycle, airplane, bus, train, truck, boat, traffic light, 
fire hydrant, stop sign, parking meter, bench, bird, cat, dog, horse, sheep, cow, 
elephant, bear, zebra, giraffe, backpack, umbrella, handbag, tie, suitcase, frisbee, 
skis, snowboard, sports ball, kite, baseball bat, baseball glove, skateboard, surfboard, tennis racket, bottle, 
wine glass, cup, fork, knife, spoon, bowl, banana, apple, sandwich, orange, 
broccoli, carrot, hot dog, pizza, donut, cake, chair, couch, potted plant, bed, 
dining table, toilet, tv, laptop, mouse, remote, keyboard, cell phone, microwave, oven, 
toaster, sink, refrigerator, book, clock, vase, scissors, teddy bear, hair drier, toothbrush, 
banner, blanket, bridge, cardboard, counter, curtain, door-stuff, flower, fruit, gravel, 
house, light, mirror-stuff, net, pillow, railroad, roof, shelf, stairs, tent, 
towel, window-blind, window-other, tree-merged, fence-merged, cabinet-merged, mountain-merged, dirt-merged, paper-merged, food-other-merged, 
building-other-merged, rock-merged, rug-merged, trash can, clothes, shadow. 
(2) “background”:
background, floor-wood, platform, playingfield, river, road, sand, sea, snow, wall-brick, wall-stone, wall-tile, wall-wood, water-other, ceiling-merged, sky-other-merged, table-merged, floor-other-merged, pavement-merged, grass-merged, wall-other-merged.

\begin{figure*}[t]
\centering
\includegraphics[width=1.0\linewidth]{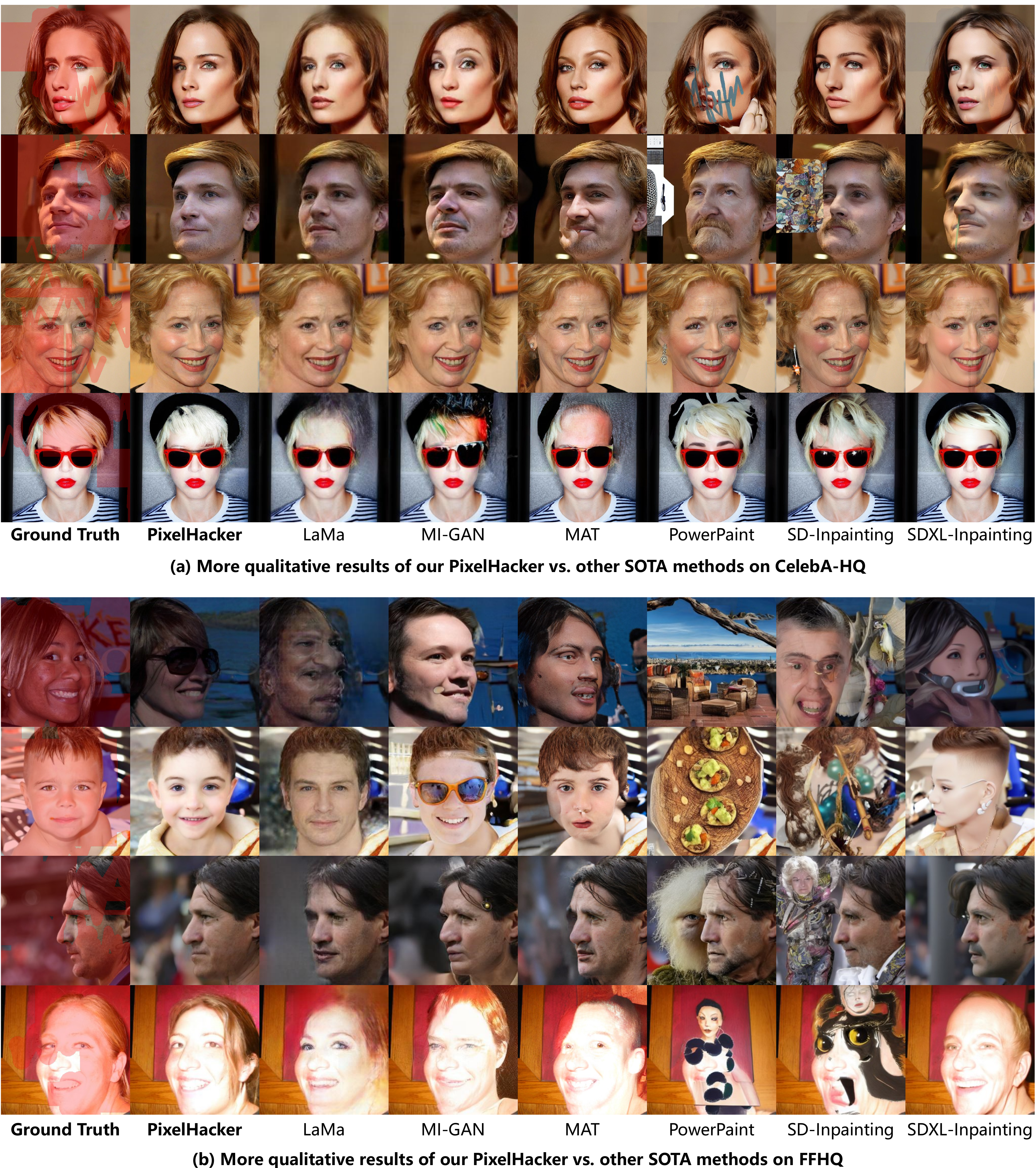}
\vspace{-6mm}
\caption{\textbf{More qualitative results of our PixelHacker vs. other SOTA methods on CelebA-HQ \cite{karras2018celebahq} and FFHQ \cite{karras2018stylegan_ffhq}.} PixelHacker preserves highly detailed facial textures while avoiding perceptible color discrepancies or artifacts, producing more realistic results than other methods and demonstrating strong adaptability to complex scene structures and challenging lighting conditions.}
\vspace{-3mm}
\label{fig:celeba_ffhq_sup}
\end{figure*}

\paragraph{More Qualitative Results.}
We provide more qualitative results of our PixelHacker with other SOTA methods on Places2, CelebA-HQ, and FFHQ, in Fig.~\ref{fig:places_sup} and Fig.~\ref{fig:celeba_ffhq_sup}.

\paragraph{Failure Cases.}
We provide some failure cases as shown in Fig.~\ref{fig:badcase}. PixelHacker works imperfectly on these examples, but is still significantly better than other methods.

\end{document}